\documentclass{article}

\usepackage[preprint]{neurips_2026}

\usepackage[utf8]{inputenc} 
\usepackage[T1]{fontenc}    
\usepackage{hyperref}       
\usepackage{url}            
\usepackage{booktabs}       
\usepackage{amsfonts}       
\usepackage{nicefrac}       
\usepackage{microtype}      
\usepackage{xcolor}         
 \usepackage{amsmath}
\usepackage{graphicx}
\usepackage{multirow}
\title{
Smooth Neural Point Processes via B-Splines}

\author{%
  Michele Bellomo
    \\
  Politecnico di Milano\\
  Milan, Italy \\
  \texttt{michele.bellomo@polimi.it} \\
  \And
  Riccardo Ramaschi \\
  Politecnico di Milano \\
  Milan, Italy \\
  \AND
  Alberto Dolara \\
  Politecnico di Milano\\
  Milan, Italy \\
  \And
  Tomaso Aste \\
  University College London \\
  London, UK \\
}

\begin{document}

\maketitle

\begin{abstract}
Temporal point processes (TPPs) provide a general and flexible framework for modeling sequences of events in continuous time. Neural networks have been successfully employed to model TPPs in a highly expressive and data-driven way. Neural TPPs are typically trained via Maximum Likelihood Estimation (MLE) 
by minimizing the negative log-likelihood (NLL), which depends on both the conditional intensity function (CIF) and its integral over time, the 
compensator. 
Recent neural TPP approaches enable exact evaluation of the NLL without numerical integration. However, these methods typically model the compensator rather than the CIF directly, impose constraints on the neural network architecture, and are computationally expensive during training, as event contributions to the NLL are evaluated sequentially rather than in parallel.
In this work, we propose a novel neural TPP model that directly parametrizes the CIF as a non-negative combination of B-spline basis functions, whose coefficients are predicted by a neural network. This formulation enables exact evaluation of the NLL, preserves full flexibility in the neural architecture, allows efficient parallelization during training, and naturally supports CIF smoothness regularization 
through the integrated squared second derivative.
Experiments on both synthetic and real-world datasets show improved computational efficiency and predictive accuracy compared to the 
reference neural TPP baseline.
\end{abstract}

\section{Introduction}
Temporal point processes (TPPs) \cite{daley} provide a general and flexible framework for modeling sequences of events in continuous time. Event sequences are common in many disciplines, such as seismology \cite{ogata2}, biology \cite{truccolo2005point}, epidemiology \cite{diggle2013spatial-temporal}, social sciences \cite{mohler2011crime}, and finance \cite{bacry, hawkes_finance, bellomo2025hawkes}.

%
A central object in the theory of TPPs is the conditional intensity function (CIF), which describes the instantaneous expected rate of events given the past history. 

Classical models such as Poisson \cite{bellomo2025hairdresser} 
and Hawkes \cite{hawkes1, laub2021} processes provide parametric forms for the CIF. These models are attractive due to their interpretability and tractability, but they may be too restrictive to capture complex temporal dependencies observed in real-world event streams. 

%
A straightforward extension of classical parametric models is to adopt a black-box approach and rely on neural networks to learn complex temporal dependencies directly from data. Despite its flexibility, this approach presents important computational challenges. Neural TPPs are typically trained via Maximum Likelihood Estimation (MLE)  by minimizing the negative log-likelihood (NLL), which depends on both the CIF and its integral over time, 
the compensator. While evaluating the CIF is straightforward, computing the integral term can be difficult when the intensity is represented by a neural network.
Some approaches \cite{jing2017neural, mei2017neural} address this issue by approximating the integral term via numerical quadrature. However, this solution is computationally expensive and may introduce instability during training. 

Omi et al. \cite{omi2019fully} proposed to use a network to model the compensator 
instead of the CIF, and then recover the CIF via automatic differentiation \cite{AD}. While this approach effectively avoids numerical integration, it requires enforcing monotonicity of the compensator 
through architectural constraints \cite{sill1997monotonic, chilinski2020neural}. These constraints limit the flexibility of the model and, importantly, prevent parallel evaluation, 
since the intensity of every time point is computed sequentially. As a result, training and inference are slow, making the approach infeasible in data-intensive settings. 

In this work, we propose a novel neural TPP model that directly parametrizes the CIF using a non-negative combination of B-spline basis functions, whose coefficients are predicted by a neural network. This formulation enables exact and efficient evaluation of the NLL, eliminating the need for numerical integration. At the same time, it preserves full flexibility in the choice of the neural architecture, as no structural constraints are required to ensure validity of the model. Importantly, the intensity of the time points can be evaluated in parallel over time, leading to significantly improved computational efficiency compared to existing approaches.
Finally, the proposed representation naturally enables smoothness regularization of the CIF via penalization of the integrated squared second derivative, providing additional control over the learned dynamics and better generalization.

\section{Background}

\subsection{Temporal Point Processes (TPPs)}
Intuitively, we can think of a point process $(T_n)_{n \geq 1}$ as an increasing sequence of random times
\[0 < T_1 < T_2 < \dots \; \;\text{.}\]
These random variables represent the occurrence (or arrival) times of events.

A point process can be associated with a counting process $N_t$, a stochastic process that counts the number of arrivals up to time 
$t$
\[
N_t = \sum_{n \geq 1} \mathbf{1}_{\{T_n\leq t\}} \;\text{.}
\]

An important quantity for a point process is the conditional intensity function (CIF), which
represents the instantaneous rate of occurrence of events at time
$t$, given the history up to time $t$ ($t$ excluded):
\[
\lambda(t \mid \mathcal{F}_t) = \lim_{\Delta t \to 0^+} \frac{\mathbb{E}(N_{t+\Delta t} - N_t \mid \mathcal{F}_t)}{\Delta t} \;\text{.}
\]
$\mathcal{F}_t$ is the filtration of the process, which represents the information available up to
time $t$. In the following, we will use the simplified notation $\lambda(t):=\lambda(t \mid \mathcal{F}_t)$, that omits the explicit dependence of the CIF on the filtration.

\subsection{Maximum Likelihood Estimation (MLE)}
The most general approach for fitting the parameters of a point process is Maximum Likelihood Estimation (MLE), which consists in maximizing the likelihood of the observed sequence of events under the assumed model.
The general form of the likelihood for point processes on the time period \( [0, T] \) is
\[
L = \left[ \prod_{i=1}^{N_T} \lambda(T_i) \right] e^{-\Lambda(T)}
\]
where $\Lambda(T) = \int_{0}^{T} \lambda(s) \, ds$ is the compensator of the process. 

For numerical reasons, it is usually preferable to maximize the log-likelihood 
\[
\ell = \left(\sum_{i=1}^{N_T} \ln(\lambda(T_i)) \right) - \Lambda(T) 
\]
or, alternatively, to minimize the negative log-likelihood (NLL), i.e., the log-likelihood with the sign reversed. 

Often, it is more convenient to work with inter-event times $\tau_i=T_i-T_{i-1}$ instead of absolute times $T_i$. In this case, the log-likelihood can be written as 
\[
\ell = \sum_{i=1}^{N_T} \left(\ln(\phi(\tau_i)) - \Phi(\tau_i) \right) 
\]
where $\phi(\tau_i)=\lambda(T_i)$ is the CIF reparameterized in terms of the time elapsed since the last event, and $\Phi(\tau_i) = \int_{0}^{\tau_i} \phi(s) \, ds$ is the relative integral. 

\subsection{Random Time Change Theorem}
\label{sec:RTCT}
The Random Time Change Theorem states that, given an increasing sequence of time points $\{T_1, T_2, \dots\}$ and a TPP with compensator $\Lambda(\cdot)$, the transformed sequence $\ \{\Lambda(T_1), \Lambda(T_2), \dots\}$ is a realisation of a unit rate Poisson process if and only if the original sequence $\{T_1, T_2, \dots\}$ is a realisation 
of the TPP defined by $\Lambda(\cdot)$. 


By the Random Time Change Theorem, when the observed event times of a TPP are transformed through its compensator, the resulting sequence follows a unit-rate Poisson process. A key property of a unit-rate Poisson process is that its inter-arrival times are independent and exponentially distributed with rate $1$, i.e.,
$
\Phi(\tau_i) = \Lambda(T_i)-\Lambda(T_{i-1}) \sim \mathrm{Exp}(1)$, with $\Phi(\tau_i)$ being the cumulative integral of the CIF reparameterized in terms of the time elapsed since the last event. 

Since an $\mathrm{Exp}(1)$ distribution has median $\log(2)$, a median-based estimator $\hat{\tau}_{i+1}$ for the waiting time until the next event can be obtained by finding $\hat{\tau}_{i+1}$  such that 
\begin{equation}
    \label{eq:median_pred}
    \Phi(\hat{\tau}_{i+1})=\log(2) \;\text{.}
\end{equation}

\subsection{B-splines}

Splines are piecewise polynomial functions defined on a partition of an interval \cite{wood2017generalized}. A spline of degree $p$ with knots $\{x_j\}$ is a function that is a polynomial of degree $p$ on each subinterval $[x_j, x_{j+1})$ and has continuous derivatives up to order $p-1$ at the knots. 

Spline spaces are dense in a wide class of function spaces, and therefore provide flexible approximations of target functions. In particular, by increasing the number of knots, spline functions can approximate arbitrarily well any target function under mild regularity conditions.

A convenient way to represent splines is through a B-spline basis \cite{nonparametric_1994}. 
Any spline function $s(x)$ can be expressed as
\begin{equation*}
s(x) = \sum_{k=1}^K w_k B_k(x)
\end{equation*}
where $w_k \in \mathbb{R}$ are coefficients and $\{B_k(x)\}_{k=1}^K$ are the non-negative B-spline basis functions. The B-spline basis has local support, meaning that each basis function is nonzero only on a limited number of adjacent knot intervals, leading to sparse and numerically stable representations. Moreover, since B-spline basis functions are non-negative, imposing non-negativity constraints on the coefficients $w_k \ge 0$ is sufficient to ensure that the resulting spline $s(x)$ is non-negative for all $x$.

Spline functions of degree $p=3$, known as cubic splines, play a particularly important role. It can be shown that, among all 
functions defined on an interval $[a,b]$ that interpolate a set of points $\{(x_i, y_i)\}$, cubic splines are the smoothest in the sense of minimizing the roughness
\begin{equation}
\label{eq:penalization}
\mathcal{P}(s) = \int_a^b (s''(x))^2 dx.
\end{equation}

In the context of nonparametric regression, cubic splines arise as the exact solution to the functional optimization problem
\begin{equation}
\label{eq:regression}
\min_{s} \ \sum_{i=1}^n \left(y_i - s(x_i)\right)^2 + \alpha \mathcal{P}(s)
\end{equation}
where $\alpha > 0$ controls the strength of the regularization.

The roughness penalty $\mathcal{P}(s)$ can be computed very efficiently. Indeed, 
it admits a quadratic form
\begin{equation}
\label{eq:penalization_calculation}
\mathcal{P}(s) = \mathbf{w}^\top \mathbf{R} \mathbf{w},
\end{equation}
where $\mathbf{w}$ is the vector of spline coefficients and $\mathbf{R}$ is a positive semidefinite matrix with entries $R_{ij} = \int_a^b B_i''(x) B_j''(x)\, dx$ that depends only on the chosen basis. Importantly, $\mathbf{R}$ can be computed once in advance, and its structure is typically sparse due to the local support of B-spline basis functions, enabling efficient evaluation of the penalty during optimization.

\section{Methodology}

\subsection{Proposed model}
We propose to model the CIF $\phi(\tau)$ on the inter-arrival times $\tau_i$ as a (cubic) B-spline function
\begin{equation*}
\phi(\tau) = \sum_{k=1}^K w_k B_k(\tau). 
\end{equation*}
The coefficients $w_k$ are predicted by a generic neural network that takes as input the history of past events and possibly 
additional contextual information.
Since B-spline basis functions are non-negative, non-negativity of the CIF is guaranteed by construction by imposing $w_k \geq 0$ through suitable output activations of the neural network, for example softplus.

A key advantage of this parameterization is that the integral of the CIF, required for the NLL evaluation, can be computed in closed form as 
\begin{equation*}
\Phi(\tau) = \int_{0}^{\tau} \phi(s) \, ds = \sum_{k=1}^K w_k I_k(\tau),
\end{equation*}
where $I_k(\tau)$ are the integrated basis functions, which can be precomputed exactly before training.

Knots $\{k_j\}$ can be distributed uniformly over the domain of inter-arrival times, or chosen according to empirical quantiles of the observed inter-arrival time distribution. To avoid oscillations in the learned CIF, one can either use a limited number of knots $K$, or increase $K$ and add the roughness penalty $\mathcal{P}$ introduced in Equation \ref{eq:penalization} to the NLL in the loss function

\begin{equation}
    \label{eq:loss}
    -\sum_{i=1}^{N_T} \left(\ln(\phi(\tau_i)) - \Phi(\tau_i)\right) + \alpha \mathcal{P}(\phi) .
\end{equation}
While no optimality result like the regression problem in Equation \ref{eq:regression} holds in the context of TPPs, penalizing the roughness of the CIF provides a principled way to control local fluctuations. 
Moreover, Equation \ref{eq:penalization_calculation} for the efficient computation of the penalization makes 
its inclusion during training essentially cost-free.

\subsection{Multivariate extension}
The proposed formulation naturally extends to multivariate TPPs with multiple event types $m \in \{1, \dots, M\}$, each associated with its own conditional intensity function $\lambda_m(t)$. 

In this setting, each conditional intensity function is modeled as
\begin{equation*}
\phi_m(\tau) = \sum_{k=1}^{K_m} w_{m,k} \, B_{m,k}(\tau),
\end{equation*}
where $\{B_{m,k}(\tau)\}_{k=1}^{K_m}$ denotes the B-spline basis associated with event type $m$, and $w_{m,k} > 0$ are the corresponding coefficients predicted by the neural network.

The extension is immediate, as it only requires increasing the dimensionality of the network output to match the total number $\sum_{m=1}^{M} K_m$ of spline coefficients across all event types.


The roughness penalty naturally generalizes as a weighted sum over event types
\begin{equation*}
P = \sum_{m=1}^{M} \alpha_m \, \mathbf{w}_m^\top \mathbf{R}_m \, \mathbf{w}_m,
\end{equation*}
where $\mathbf{w}_m = (w_{m,1}, \dots, w_{m,K_m})^\top$ and $\mathbf{R}_m$ is the precomputed roughness matrix associated with the basis of type $m$. The coefficients $\alpha_m \geq 0$ control the strength of the smoothness regularization for each event type, allowing different levels of roughness penalization across the corresponding intensity functions.

For simplicity, one can also choose to use a shared temporal basis across all event types, instead of defining a distinct spline basis with different knot placements and numbers for each type.

\subsection{Goodness of fit}
\label{sec:diagnostics}
Similarly to \cite{omi2019fully}, we evaluate the goodness of fit of the proposed model on the test data using 
the mean absolute error (MAE) computed on the median-based estimator $\hat{\tau}_{i+1}$ for the next inter-arrival time described in Section \ref{sec:RTCT} with respect to the true observed $\tau_{i+1}$. 

Our model allows 
to obtain the estimator $\hat{\tau}_{i+1}$ in 
a very efficient way. In fact, instead of evaluating the CIF and the compensator in a unique 
future time as in \cite{omi2019fully}, with a single forward pass our model outputs the spline coefficients from which the entire future trajectories of both the CIF and the compensator are obtained. This enables solving Equation \ref{eq:median_pred} very efficiently, as no additional forward passes of the model are required during the iterations of the root-finding algorithm. Given the monotonicity of the compensator, a simple yet effective approach to solve Equation \ref{eq:median_pred} is the bisection method.

In case of a multivariate TPP, the median-based estimator $\hat{\tau}_{i+1}$ for the next inter-arrival time is computed using the total CIF $\phi(\tau) = \sum_{m=1}^{M} \phi_m(\tau)$ and the corresponding total compensator $\Phi(\tau) = \sum_{m=1}^{M} \Phi_m(\tau)$. 
The estimated next event type 
can be determined by 
\[
\hat{m}_{i+1}
=
{\arg\max}_{m \in \{1,\dots,M\}} \;
\phi_m(\hat{\tau}_{i+1}).
\]

\section{Experiments}

\subsection{Datasets}
\label{sec:datasets}
We test our model on the same synthetic and real datasets used in \cite{omi2019fully}, whose model will be used as the baseline for comparison.

Each synthetic dataset consists of a sequence of $100000$ time points, generated by different TPP models. 
Every sequence is then split into train
and test sets, using a standard 80\%/20\% division. The TPPs considered are:
\begin{itemize}
    \item a stationary Poisson process with $\lambda(t)=1$;
    \item a non-stationary Poisson process with $\lambda(t)=0.99\cdot sin(\frac{2\pi t}{2000})+1$;
    \item a stationary Renewal process, in which the inter-event intervals $\tau_i$ are independent and identically distributed according to a log-normal probability distribution with mean \(\mu = 1.0\) and standard deviation \(\sigma = 6.0\);
    \item a non-stationary Renewal process, whose time points $T_i$ are obtained by first generating independent and identically distributed inter-event times $\tau'_i$ from a stationary Renewal process (Gamma distribution with mean \(\mu = 1.0\) and standard deviation \(\sigma = 0.5\)), and then rescaling the time by inverting the formula $\tau'_i = \int_{T_{i-1}}^{T_i} r(s) \, ds$, where \(r(t)\) is a non-negative trend function defined as \(
r(t) = 0.99 \, \sin\Big(\frac{2\pi t}{20000}\Big) + 1 \);
\item a Self-correcting process, with $\lambda(t)=\exp(t-\sum_{t_i<t}1)$;
\item a Hawkes process with single exponential kernel  \(
\lambda(t) = \mu + \sum_{T_i < t} \alpha\, \beta \, 
\exp\Big\{ - \beta (t - T_i) \Big\}
\) and parameters $\mu=0.2$, $\alpha=0.8$, $\beta=1$;
\item a Hawkes process with multiple exponential kernel \[
\lambda(t) = \mu + \sum_{T_i < t} \sum_{j=1}^{M} \alpha_j \, \beta_j \, 
\exp\Big\{ - \beta_j (t - T_i) \Big\}\]
and parameters $M=2$, $\mu=0.2$, $\alpha_1=\alpha_2=0.4$, $\beta_1=0.4$, $\beta_2=20$.

\end{itemize}

We additionally use the two publicly available real-world datasets employed in \cite{omi2019fully}. For both datasets, we apply the same preprocessing described in \cite{omi2019fully}, and time is measured in hours.
\begin{itemize}
    \item Music dataset \cite{music} 
    contains last.fm users' listening history in January 2009. 
    100 sequences from the 100 most active users are created, using the first 80\% of events in the sequences 
    for training and the last 20\% 
    for testing.
    \item Meme dataset \cite{memetracker} 
    contains popular phrases collected from several online resources. 
    50 sequences from the 50 most used phrases during August 2008 are created, using the first 40 sequences for training and the last 10 sequences for testing.
\end{itemize}



\subsection{Model architecture and training procedure}
We use the same architecture for all datasets: a minimal Transformer with embedding size 64, 4 attention heads, and a single vanilla 
decoder block as described in \cite{attention2017}, with an inner feed-forward layer dimension of 64. 
The input embedding is constructed as the sum of two components: a sinusoidal positional encoding to represent the chronological order of past events, and a learned embedding of the inter-arrival times obtained through a linear dense layer.

The CIF curve over future times is predicted by taking as input the inter-arrival times of the last 20 events.
The CIF is parametrized as a cubic B-spline function with 20 internal knots distributed at quantiles of inter-arrival training times. The network outputs the positive coefficients of the B-splines using softplus activation in the last layer. 

The network is trained minimizing the penalized loss in Equation \ref{eq:loss} using Adam optimizer with learning rate 0.001, batch size 256, and maximum training epochs 200. Early stopping is applied based on the validation set, obtained as 20\% of the training data, with patience 10 epochs, the NLL without penalization as monitored validation loss, and restoring the best weights at the end of the training.

The validation set is also used to select for each dataset the optimal value of the regularization parameter $ \alpha$, evaluated over a range of values from $10^{-3}$ to $10^{-10}$, also including the case with no regularization $\alpha=0$.
All other hyperparameters were chosen to be reasonable for the considered case studies and were not subject to any optimization.

Our implementation relies on \texttt{Keras} \cite{keras} and \texttt{KerasHub} \cite{kerashub} libraries, and experiments were run on a T4 GPU provided by the Google Colab environment.

\subsection{Computational performance evaluation}
\label{sec:computational_performance}
To ensure a fair comparison with the benchmark in \cite{omi2019fully}, the predictive accuracy results are obtained under the same training setting, using the cut sequence of last inter-event times to predict only the CIF at the next event. 

However, our spline-based neural TPP model allows parallel evaluation of the CIF over multiple future times. In particular, given a sequence of events, our model is able to return the CIF evaluated at the time of each event in a single forward pass. This significantly reduces the training and testing time, and is especially beneficial when using a Transformer architecture.

We compare the training computational cost of the CIF evaluation scheme used in \cite{omi2019fully} with the parallel multi-evaluation variant of our model. In the latter setting, the model processes longer sequences (in our experiments, of length 100) and directly outputs the CIF at each event time in a single forward pass. A warm-up period can be introduced, excluding the first n events from the loss computation, as they may not have sufficient conditioning history (in our experiments, $n=20$).
Finally, when using a Transformer architecture, it is also possible to limit the effective context by restricting attention to the most recent n past events through a custom attention mask ($n=20$ in our experiments).

\subsection{Results}
Table \ref{tab:results} shows the results of the experiments described in Section \ref{sec:diagnostics} on the datasets introduced in Section \ref{sec:datasets}. For each model, the best regularization parameter $\alpha$ is selected on the validation set, and then the MAE of the median-based estimator is computed on the test set and compared with the one reported by Omi et al. \cite{omi2019fully}. 


\begin{table}[h]
  \centering
  \caption{Time prediction experiment results.}
  \label{tab:results}
  \begin{tabular}{ccccc}
    \toprule
    Type & Dataset & MAE & Omi et al. MAE \cite{omi2019fully} \\
    \midrule
    \multirow{7}{*}{Synthetic} 
    & Stationary Poisson & \textbf{0.692} & 0.696 \\
    & Non-stationary Poisson & 0.711 & \textbf{0.710} \\
    & Stationary Renewal & 0.957 & \textbf{0.894} \\
    & Non-stationary Renewal & \textbf{0.406} & 0.414 \\
    & Self-correcting & \textbf{0.494} & 0.496 \\
    & Hawkes 1 & \textbf{0.399} & 0.848 \\
    & Hawkes 2 & \textbf{0.947} & 0.962 \\
    \midrule
    \multirow{2}{*}{Real}
    & Music & \textbf{0.183} & 0.783 \\
    & Meme & \textbf{0.135} & 0.811 \\
    \bottomrule
  \end{tabular}
\end{table}

Our model outperforms the baseline in 5 out of 7 synthetic datasets and in both real-world datasets. On the ``Hawkes 1'' dataset and on the real-world datasets, the improvement in MAE is substantial.

\begin{figure}[!h]
    \centering
    \includegraphics[width=\linewidth]{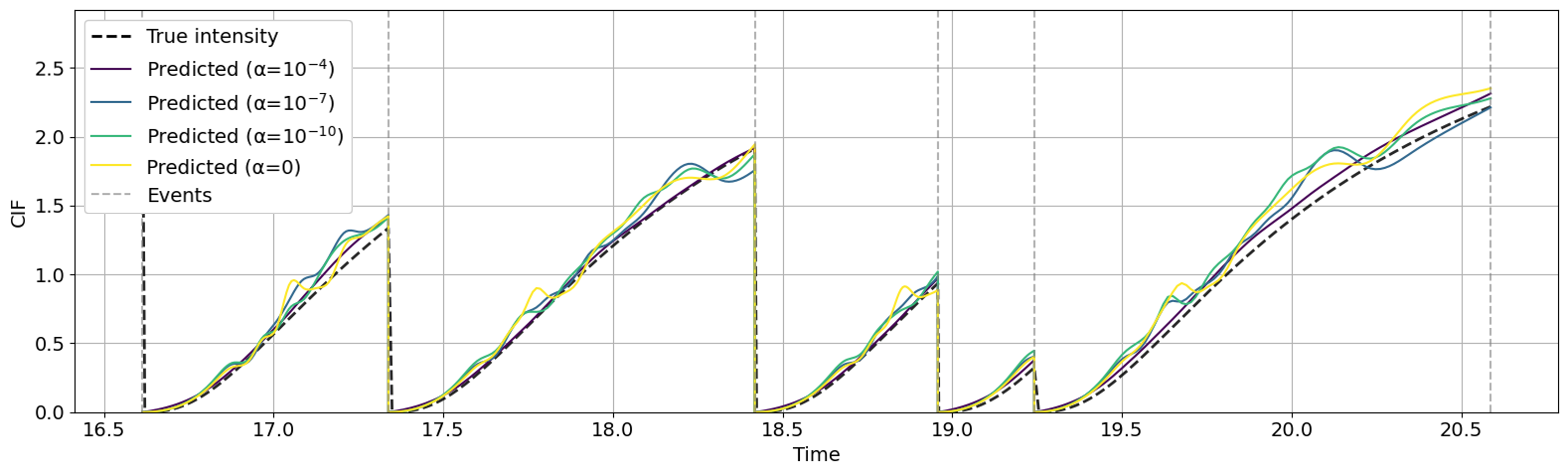}
    \caption{Regularization effect of different smoothing parameters $\alpha$ on the learned CIF from the non-stationary Renewal process.}
    \label{fig:N-Renewal}
\end{figure}

\begin{figure}[!h]
    \centering
    \includegraphics[width=\linewidth]{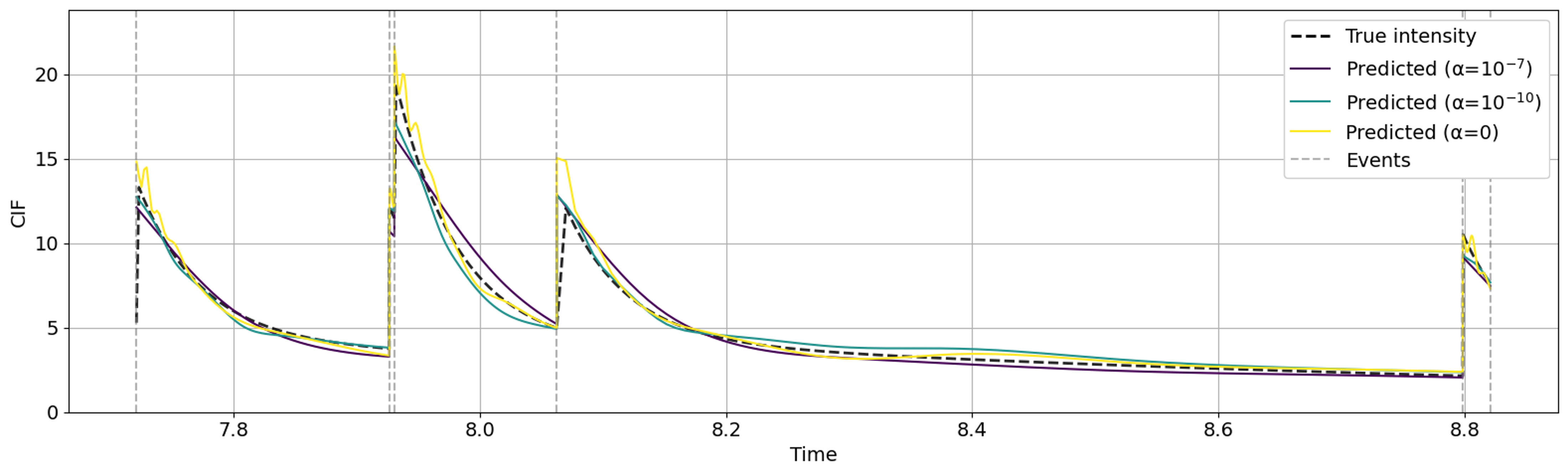}
    \caption{Regularization effect of different smoothing parameters $\alpha$ on the learned CIF from the Hawkes 2 process.}
    \label{fig:Hawkes2}
\end{figure}

In most experiments, the optimal smoothing parameter $\alpha$ was nonzero, suggesting that regularization was indeed beneficial. The effect of regularization on the learned CIF from non-stationary Renewal and Hawkes 2 processes is shown in Figures \ref{fig:N-Renewal} and \ref{fig:Hawkes2}. 

A higher value of the regularization parameter $\alpha$ reduces the fluctuations of the predicted CIF, as is evident in the non-stationary Renewal process in Figure \ref{fig:N-Renewal}. 
In both cases, values of $\alpha$ not far from the optimal one do not lead to a significant degradation in performance, indicating that the model is reasonably robust to this parameter. 

Using the T4 GPU provided by the Google Colab environment, one epoch with the standard training setting takes on average 1750 ms on synthetic datasets, whereas with our parallel multi-evaluation approach described in Section \ref{sec:computational_performance}, an epoch takes on average 150 ms, corresponding to a speed-up of approximately 12×.

\section{Discussion}
In this work, we introduced a novel neural TPP model that directly parametrizes the CIF using a non-negative combination
of B-spline basis functions, whose coefficients are predicted by a generic neural network. This method preserves exact evaluation of the NLL, but, unlike previous approaches such as \cite{omi2019fully}, it allows full flexibility in the choice of the neural architecture and, thanks to the extensive parallelization, enables fast training on large datasets. 
Moreover, the smoothing regularization through the integrated squared second derivative prevents the oscillatory behavior typically associated with splines and leads to improved generalization. 

We evaluated the proposed approach on both synthetic and real-world datasets commonly used as benchmarks for neural TPPs. The results show that parametrizing the CIF with B-splines does not degrade predictive accuracy, and in fact leads to a significant improvement, in particular on the real-world datasets.  

In addition to improved predictive accuracy, the proposed model is computationally more efficient thanks to its flexibility, enabling the use of architectures such as Transformers and parallel processing across sequence events, unlike the sequential computation required by the recurrent neural network in \cite{omi2019fully}. 
This advantage becomes substantial when adopting the multi-evaluation approach introduced in Section \ref{sec:computational_performance}, which allows the CIF to be evaluated for all events in the sequences in a single forward pass. This resulted in an empirical training speedup of approximately 12× on the synthetic datasets considered in our experiments.

%

\section{Limitations and future work}
The proposed model introduces an additional hyperparameter, namely the regularization coefficient $\alpha$. While the model is significantly more efficient due to the use of a Transformer and the parallel multi-evaluation strategy, tuning $\alpha$ may partially offset these computational gains. In future work, we aim to investigate more efficient and theoretically grounded strategies for the automatic tuning of this parameter. 

The proposed methodology shows promising results, but it has so far been evaluated only on
datasets of moderate size and using minimal architectures. In future work, we plan to assess the proposed method on in-depth real-world case studies involving large-scale datasets 
with multiple event types, where other neural TPP approaches may become infeasible and where the computational efficiency and scalability of the proposed framework can be more fully appreciated.


\bibliographystyle{unsrt}  
\bibliography{references}  







\end{document}